\documentclass[sigconf]{acmart}

\AtBeginDocument{%
  \providecommand\BibTeX{{%
    \normalfont B\kern-0.5em{\scshape i\kern-0.25em b}\kern-0.8em\TeX}}}

\usepackage[utf8]{inputenc} % allow utf-8 input
\usepackage[T1]{fontenc}    % use 8-bit T1 fonts
\usepackage{hyperref}       % hyperlinks
\usepackage{url}            % simple URL typesetting
\usepackage{booktabs}       % professional-quality tables
\usepackage{amsfonts}       % blackboard math symbols
\usepackage{nicefrac}       % compact symbols for 1/2, etc.
\usepackage{microtype}      % microtypography
\usepackage{amssymb}
\usepackage{amsmath}
\usepackage{subfigure}
\usepackage{lipsum,multicol}

\usepackage{multirow}
\usepackage{array}
\newcolumntype{L}[1]{>{\raggedright\let\newline\\\arraybackslash\hspace{0pt}}m{#1}}
\newcolumntype{C}[1]{>{\centering\let\newline\\\arraybackslash\hspace{0pt}}m{#1}}
\newcolumntype{R}[1]{>{\raggedleft\let\newline\\\arraybackslash\hspace{0pt}}m{#1}}

\usepackage[font=small,labelfont=bf]{caption}

\usepackage{tikz}
\usepackage{graphicx}
\usetikzlibrary{shapes.misc, positioning}
\usepackage{xcolor}
\usepackage{hyphenat}
\usepackage{enumitem,kantlipsum}
\definecolor{c1}{cmyk}{0.0, 0.67, 0.81, 0}
\newcommand{\bx}{\mathbf{x}}
\newcommand{\by}{\mathbf{y}}

\newcommand{\bh}{\mathbf{h}}
\newcommand{\bq}{\mathbf{q}}
\newcommand{\bk}{\mathbf{k}}
\newcommand{\bv}{\mathbf{v}}
\newcommand{\bd}{\mathbf{d}}
\newcommand{\be}{\mathbf{e}}
\newcommand{\bff}{\mathbf{f}}

\newcommand{\bc}{\mathbf{c}}

\newcommand{\BR}{\mathbb{R}}
\newcommand{\BN}{\mathbb{N}}

\newcommand{\rom}[1]{\uppercase\expandafter{\romannumeral #1\relax}}
\title{Context-Aware Attentive Knowledge Tracing}

\copyrightyear{2020} 
\acmYear{2020} 
\setcopyright{acmcopyright}\acmConference[KDD '20]{Proceedings of the 26th ACM SIGKDD Conference on Knowledge Discovery and Data Mining USB Stick}{August 23--27, 2020}{Virtual Event, USA}
\acmBooktitle{Proceedings of the 26th ACM SIGKDD Conference on Knowledge Discovery and Data Mining USB Stick (KDD '20), August 23--27, 2020, Virtual Event, USA}
\acmPrice{15.00}
\acmDOI{10.1145/3394486.3403282}
\acmISBN{978-1-4503-7998-4/20/08}

\begin{document}
\fancyhead{}

%\title{The Name of the Title is Hope}

\author{Aritra Ghosh}
\affiliation{%
  \institution{University of Massachusetts Amherst}
  \city{Amherst}
  \state{MA}
}
\email{arighosh@cs.umass.edu}

\author{Neil Heffernan}
\affiliation{%
  \institution{Worcester Polytechnic Institute}
  \city{Worcester}
  \country{MA}}
\email{nth@wpi.edu}

\author{Andrew S.\ Lan}
\affiliation{%
  \institution{University of Massachusetts Amherst}
  \city{Amherst}
  \state{MA}
}
\email{andrewlan@cs.umass.edu}

\renewcommand{\shortauthors}{Ghosh, Heffernan, Lan}

\begin{abstract}
Knowledge tracing (KT) refers to the problem of predicting future learner performance given their past performance in educational applications. Recent developments in KT using flexible deep neural network-based models excel at this task. However, these models often offer limited interpretability, thus making them insufficient for personalized learning, which requires using interpretable feedback and actionable recommendations to help learners achieve better learning outcomes.
In this paper, we propose attentive knowledge tracing (AKT), which couples flexible attention-based neural network models with a series of novel, interpretable model components inspired by cognitive and psychometric models. 
AKT uses a novel monotonic attention mechanism that relates a learner's future responses to assessment questions to their past responses; attention weights are computed using exponential decay and a context-aware relative distance measure, in addition to the similarity between questions.
Moreover, we use the Rasch model to regularize the concept and question embeddings; these embeddings are able to capture individual differences among questions on the same concept without using an excessive number of parameters. 
We conduct experiments on several real-world benchmark datasets and show that AKT outperforms existing KT methods (by up to $6\%$ in AUC in some cases) on predicting future learner responses. 
We also conduct several case studies and show that AKT exhibits excellent interpretability and thus has potential for automated feedback and personalization in real-world educational settings.
\end{abstract}
\maketitle
	
	\section{Introduction}
\label{sec:intro}
Recent advances in data analytics and intelligent tutoring systems \cite{bevbook} have enabled the collection and analysis of large-scale learner data; these advances hint at the potential of \emph{personalized} learning at large scale, by automatically providing personalized feedback \cite{piechfeedback} and learning activity recommendations \cite{banditslan} to each learner by analyzing data from their learning history. 

A key problem in learner data analysis is to predict future learner performance (their responses to assessment questions), given their past performance, which is referred to as the knowledge tracing (KT) problem \cite{kt}. Over the last 30 years, numerous methods for solving the KT problem were developed based on two common assumptions: i) a learner's past performance can be summarized by a set of variables representing their current latent \emph{knowledge level} on a set of concepts/skills/knowledge components, and ii) a learner's future performance can be predicted using their current latent concept knowledge levels. Concretely, let $t$ denote a set of discrete time indices, we have the following generic model for a learner's knowledge and performance
\begin{align*}
r_t \sim f(h_t), \quad h_t \sim g(h_{t-1}),
\end{align*}
where $r_t \in \{0,1\}$ denotes the learner's graded response to an assessment question at time step~$t$, which is usually binary-valued ($1$ corresponds to a correct response and $0$ corresponds to an incorrect one) and is observed. The latent variable $h_t$ denotes the learner's current knowledge level and is not observed. $f(\cdot)$ and $g(\cdot)$ are functions that characterize how learner knowledge dictate their responses and how it evolves; they are sometimes referred to as the response model and the knowledge evolution model, respectively. 

Earlier developments in KT methods before 2010 can be divided into two classes. 
The first class centered around the Bayesian knowledge tracing (BKT) method \cite{pardoskt,bkt} where knowledge ($h_t$) is a binary-valued scalar that characterizes whether or not a learner masters the (single) concept covered by a question. Since the response ($r_t$) is also binary-valued, the response and knowledge evolution models are simply noisy binary channels, parameterized by the guessing, slipping, learning, and forgetting probabilities. 
The second class centered around item response theory (IRT) models \cite{lordirt} and use these models (especially sigmoidal link functions) as the response model $f(\cdot)$; learner knowledge level is then modeled as real-valued vectors ($\mathbf{h}_t$) for questions that cover multiple concepts. Among these methods, the SPARFA-Trace method \cite{tracekdd} used a simple affine transformation model as the explicit knowledge evolution model $g(\cdot)$. Other methods, e.g., additive factor models \cite{lfa}, performance factor analysis \cite{pfa}, the difficulty, ability, and student history (DASH) model \cite{dash}, and a few recent methods including knowledge factorization machines \cite{kfm} and an extension to the DASH model, the DAS3H model \cite{das3h}, used \emph{hand-crafted features} such as the number of previous attempts, successes, and failures on each concept in their knowledge evolution model. 
Methods in both classes rely on expert labels to associate questions to concepts, resulting in excellent interpretability since they can effectively estimate the knowledge level of each learner on expert-defined concepts.

Recent developments in KT centered around using more sophisticated and flexible models to fully exploit the information contained in large-scale learner response datasets. 
The deep knowledge tracing (DKT) method \cite{dkt} was the first method to explore the use of (possibly deep) neural networks for KT by using long short-term memory (LSTM) networks \cite{lstm} as the knowledge evolution model $g(\cdot)$. Since LSTM units are nonlinear, complex functions, they are more flexible than affine transformations and more capable of capturing nuances in real data. 

The dynamic key-value memory networks (DKVMN) method extended DKT by using an external memory matrix ($\mathbf{H}_t$) to characterize learner knowledge \cite{dkvmn}. The matrix is separated into two parts: a static, ``key'' matrix that contains the fixed representation of each concept, and a dynamic, ``value'' matrix that contains the evolving knowledge levels of each learner on each concept. DKVMN also uses separate ``read'' and ``write'' processes on this external matrix for the response and knowledge evolution models; these processes make it even more flexible than DKT. 
DKT and KVMN reported state-of-the-art performance in predicting future learner performance \cite{howdeep} and have been the benchmark for new KT methods. 

The self-attentive knowledge tracing (SAKT) method \cite{sakt} is the first method to use attention mechanisms in the context of KT. Attention mechanisms are more flexible than recurrent and memory-based neural networks and have demonstrated superior performance in natural language processing tasks. 
The basic setup of SAKT has many similarities to the Transformer model \cite{transformer}, an effective model for many sequence-to-sequence prediction tasks.
However, we observe that SAKT does not outperform DKT and DKVMN in our experiments; see Section~\ref{sec:experiments} for details. 
Possible reasons for this include i) unlike in language tasks where strong long-distance dependencies between words are more common, the dependence of future learner performance on the past is likely restricted to a much shorter window, and ii) the sizes of learner response datasets are several magnitudes lower than natural language datasets and are less likely to benefit from highly flexible and large-scale attention models. 

More importantly, no existing KT method truly excels at both future performance prediction and interpretability. Early KT methods exhibit excellent interpretability but do not provide state-of-the-art performance on future learner performance prediction. Recent KT methods based on deep learning excel at that but offer limited interpretability.  Therefore, these KT methods do not fully meet the needs of personalized learning, which requires not only accurate performance prediction but also the ability to provide automated, interpretable feedback and actionable recommendations to help learners achieve better learning outcomes.

\subsection{Contributions}
For the task of predicting the learner's response to the current question, we propose the attentive knowledge tracing (AKT) method, which uses a series of attention networks to draw connections between this question and every question the learner has responded to in the past. We summarize our key innovations below:
\begin{enumerate}[leftmargin=*]
\item Contrary to existing attention methods that use raw question and response embeddings, we put raw embeddings into context and use \emph{context-aware representations} of past questions and responses by taking a learner's entire practice history into account. 
\item Inspired by cognitive science findings on the mechanisms of forgetting, we propose a novel \emph{monotonic attention mechanism} that uses an exponential decay curve to down weight the importance of questions in the distant past. We also develop a context-aware measure to characterize the time distance between questions a learner has responded to in the past. 
\item Leveraging the Rasch model, a simple and interpretable IRT model, we use a series of \emph{Rasch model-based embeddings} to capture individual differences among questions without introducing an excessive amount of model parameters. 
\end{enumerate}
We conduct a series of experiments on several benchmark real-world educational datasets comparing AKT to state-of-the-art KT methods. Our results show that AKT (sometimes significantly) outperforms other KT methods in predicting future learner performance. Further, we perform ablation studies on each of the key AKT model components to demonstrate their value. 
We also perform several case studies to show that AKT exhibits excellent interpretability and has the potential for automated feedback and practice question recommendation, both key requirements of personalized learning\footnote{Source code and datasets will be available at \href{https://github.com/arghosh/AKT}{https://github.com/arghosh/AKT}}.

	\section{Knowledge Tracing Problem Setup}
Each learner's performance record consists of a sequence of questions and responses at each discrete time step. 
For learner $i$ at time step $t$, we denote the combination of the question that they answered, the concept this question covers, and their graded response as a tuple, $(q_t^i, c_t^i,  r_t^i)$, where $q_t^i\in \BN^{+}$ is the question index, $c_t^i\in \BN^{+}$ is the concept index, and $r_t^i\in \{0,1\}$ is the response. Under this notation, $(q_t^i, c_t^i, 1)$ means learner $i$ responded to question $q_t^i$ on concept $c_t^i$ correctly at time $t$. 
We note that this setup is different from some prior works on deep knowledge tracing that often ignore the question index and summarize learner performance as $(c_t^i,  r_t^i)$. This choice was made to avoid overparameterization; see Section~\ref{sec:rasch} for a detailed analysis. 
In the following discussions, we omit the superscript $i$ as we discuss how to predict future performance for a single learner. 
Given their past history up to time $t-1$ as $\{(q_1, c_1, r_1), \ldots, (q_{t-1}, c_{t-1}, r_{t-1})\}$, our goal is to predict their response $r_t$ to question $q_t$ on concept $c_t$ at the current time step, $t$.

\subsection{Question and Response Embeddings} 
\label{sec:emb}
Following previous work \cite{dkvmn},
we use real-valued embedding vectors $\bx_t \in \BR^{D}$ and $\by_t \in \BR^{D}$ to represent each question and each question-response pair ($q_t, r_t$), respectively. $\bx_t$ characterizes information about questions, and $\by_t$ characterizes the knowledge learners acquire by responding to questions, with two separate embeddings for correct and incorrect responses, respectively. $D$ denotes the dimension of these embeddings. 
Therefore, let $Q$ denote the number of questions, there are a total of $Q$ question embedding vectors and $2Q$ question-response embedding vectors. 
In most real-world educational settings, the question bank is considerably larger than the set of concepts and many questions are assigned to very few learners. Therefore, the majority of existing KT methods use concepts to index questions to avoid overparameterization; all questions covering the same concept are treated as a single question. In this case, $q_t = c_t$ and $Q=C$. 

\section{The AKT Method}
\begin{figure*}[tp]
    \centering
    \includegraphics[width=1.9\columnwidth]{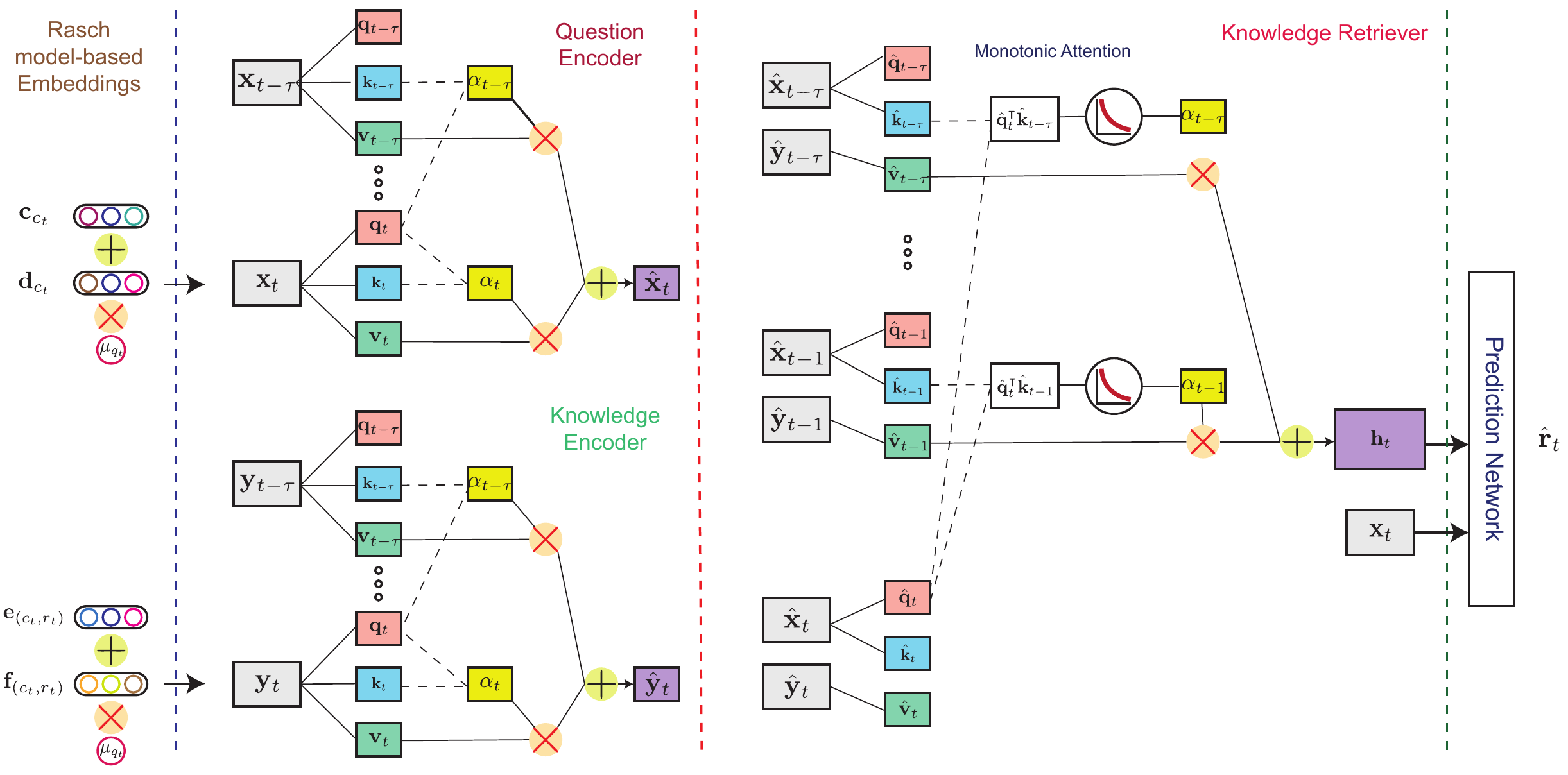}
    \caption{Overview of the AKT method. We use the Rasch model-based embeddings as raw embeddings for questions and responses. The question and knowledge encoders compute the context-aware representations of questions and responses pairs. The knowledge retriever uses these representations as input and computes the knowledge state of the learner. For simplicity, we do not show the monotonic attention mechanism in the encoders. We do not show sublayers either.}
    \label{fig:main}
\end{figure*}

The AKT method consists of four components: two self-attentive encoders, one for questions and one for knowledge acquisition, a single attention-based knowledge retriever, and a feed-forward response prediction model; Figure~\ref{fig:main} visualizes the AKT method and its connected components.

We use the two self-attentive encoders to learn context-aware representations of the questions and responses. 
We refer to the first encoder as the \emph{question encoder}, which produces modified, contextualized representations of each question, given the sequence of questions the learner has previously practiced on. 
Similarly, we refer to the second encoder as the \emph{knowledge encoder}, which produces modified, contextualized representations of the knowledge the learner acquired while responding to past questions. Alternatively, we could use raw embeddings of questions and responses similar to prior work. We found that the context-aware representation performs better in most datasets. 
We refer to the knowledge evolution model as the \emph{knowledge retriever}, which retrieves knowledge acquired in the past that is relevant to the current question using an attention mechanism. 
Finally, the response prediction model predicts the learner's response to the current question using the retrieved knowledge. 
The AKT method is motivated by three intuitions rooted in cognitive science and psychometrics; we will detail these intuitions in what follows.

\subsection{Context-aware Representations and The Knowledge Retriever} 
As introduced above, we use two encoders in our model. The question encoder takes raw question embeddings $\{\bx_1, \ldots, \bx_{t}\}$ as input and outputs a sequence of context-aware question embeddings $\{\hat{\bx}_1,\ldots$, $ \hat{\bx}_{t} \}$ using a monotonic attention mechanism (detailed in the next subsection). The context-aware embedding of each question depends on both itself and the past questions, i.e., $\hat{\bx}_{t}  =  f_{\mbox{enc}_1}$ $(\bx_{1},\ldots,\bx_{t})$. 
Similarly, the knowledge encoder takes raw question-response embeddings $\{\by_1, \ldots, \by_{t-1}\}$ as input and outputs a sequence of actual knowledge acquired $\{\hat{\by}_1,\ldots, \hat{\by}_{t-1} \}$ using the same monotonic attention mechanism. The context-aware embedding of acquired knowledge depend on the learner's response to both the current question and past questions, i.e., $\hat{\by}_{t-1}=  f_{\mbox{enc}_2}(\by_{1},\ldots,\by_{t-1})$. 

The choice of using context-aware embeddings rather than raw embeddings reflects our first intuition: \emph{the way a learner comprehends and learns while responding to a question depends on the learner}. These modified representations reflect each learner's actual comprehension of the question and the knowledge they actually acquire, given their personal response history. This model choice is motivated by the intuition that for two learners with different past response sequences, the way they understand the same question and the knowledge they acquire from practicing on it can differ.

The knowledge retriever takes the context-aware question and question-response pair embeddings $\hat{\bx}_{1:t}$ and $\hat{\by}_{1:t-1}$ as input and outputs a retrieved knowledge state $\bh_t$ for the current question. We note that in AKT, the learner's current knowledge state is also context-aware since it depends on the current question they are responding to; this model choice is different from that in most existing methods, including DKT. We also note that the knowledge retriever can only use information on the past questions, the learner's responses to them, and the representation of the current question, but not the learner's response to the current question, i.e., $\bh_{t} = f_{\mbox{kr}}(\hat{\bx}_1,\ldots, \hat{\bx}_{t}, \hat{\by}_{1},\ldots,\hat{\by}_{t-1})$. The response prediction model uses the retrieved knowledge  to predict the current response. 

\subsection{The Monotonic Attention Mechanism}
\label{sec:ma}
We use a modified, monotonic version of the scaled dot-product attention mechanism for the encoders and the knowledge retriever. We start by briefly summarizing the original scaled dot-product attention mechanism. Under this framework, each encoder and the knowledge retriever has a key, query, and value embedding layer that maps the input into output queries, keys, and values of dimension $D_q = D_k$, $D_k$, and $D_v$, respectively. 
Let $\bq_t \in \BR^{D_k \times 1}$ denote the query corresponding to the question the learner responds to at time $t$, the scaled dot-product attention values $\alpha_{t,\tau}$ are computed using the softmax function \cite{dlbook} as
\begin{align*}
\alpha_{t,\tau} = \mbox{Softmax}(\frac{\bq_t^{\intercal} \bk_\tau}{\sqrt{D_k}}) = \frac{\exp(\frac{\bq_t^{\intercal} \bk_\tau}{\sqrt{D_k}})}{\sum_{\tau'} \exp(\frac{\bq_t^{\intercal} \bk_\tau}{\sqrt{D_k}})} \in [0,1]. 
\end{align*} 
The output of the scaled dot-product attention mechanism is then given by $\textstyle\sum_{\tau} \alpha_{t,\tau} \bv_\tau \in \BR^{D_v\times 1}$. $\bk_\tau \in \BR^{D_k\times 1}$ and $\bv_\tau \in \BR^{D_v\times 1}$ denote the key and value for the question at time step $\tau$, respectively.
Depending on the specific component, the output depends either on both the past and the current ($\tau \leq t$ for the question and knowledge encoders) or only the past ($\tau < t$ for the knowledge retriever). 

\sloppy
Both encoders employ the self-attention mechanism, i.e., $\bq_t$, $\bk_t$, and $\bv_t$ are computed using the same input; the question encoder uses $\{\bx_1, \ldots,\bx_t\}$ while the knowledge encoder uses $\{\by_1, \ldots,\by_{t-1}\}$.  
The knowledge retriever, on the other hand, does not use self-attention. As shown in Fig.~\ref{fig:main}, at time step $t$, it uses $\hat{\bx}_{t}$ (the modified embedding of the current question), $\{\hat{\bx}_1,\ldots,\hat{\bx}_{t-1}\}$ (the context-aware embeddings of past questions), and $\{\hat{\by}_1,\ldots,\hat{\by}_{t-1}\}$ (the context-aware embeddings of past question-response pairs) as input to generate the query, keys, and values, respectively. We note that SAKT uses question embeddings for mapping queries whereas response embeddings for key and value mapping. 
%However, ideally model should look for similarity in questions only for computing attention scores. 
In our experiments, we have found that using question embeddings for mapping both queries and keys is much more effective.

However, this basic scaled dot-product attention mechanism is not likely going to be sufficient for KT. The reason is that learning is temporal and memories decay \cite{mcm}; a learner's performance in the distant past is not as informative as recent performance when we are predicting their response to the current question. 
Therefore, we develop a new monotonic attention mechanism that reflects our second intuition: \emph{when a learner faces a new question, past experiences i) on unrelated concepts and ii) that are from too long ago are not likely to be highly relevant.} 
Specifically, we add a multiplicative exponential decay term to the attention scores as: 
\begin{align*}
\alpha_{t,\tau} = \frac{\exp{(s_{t,\tau})}}{\sum_{\tau'} \exp{(s_{t,\tau'})}},
\end{align*}
with
\begin{align}
s_{t,\tau} &=  \frac{\exp \left(- \theta\cdot  d(t,\tau) \right) \cdot \bq_t^{\intercal} \bk_\tau }{\sqrt{D_k}}, 
\end{align}
where $\theta > 0$ is a learnable decay rate parameter and $d(t,\tau) $ is temporal distance measure between time steps $t$ and $\tau$.
In other words, the attention weights for the current question on a past question depends not only on the similarity between the corresponding query and key, but also on the relative number of time steps between them. In summary, our monotonic attention mechanism takes the basic form of an exponential decay curve over time with possible spikes at time steps when the past question is highly similar to the current question. We note that we apply exponential decay to the attention weights rather than latent knowledge, which is the common approach in existing learner models (see e.g., \cite{reddy,moz}). 

We note that there are many other possible ways to characterize the temporal dynamics of attention. First, in language tasks that attention networks excel at, the temporal dynamics can be modeled using additive positional embeddings or learnable embeddings \cite{transformer}. 
Second, in our monotonic attention mechanism, we can also parameterize the exponential decay as $s_{t,\tau} = \frac{\bq_t^{\intercal} \bk_\tau}{\sqrt{D_k}} - \theta\cdot d(t ,\tau)$. 
However, neither of these changes lead to comparable performance to our chosen model setup; we will compare AKT against its variants using positional encoding rather than monotonic attention in our experiments.

\textbf{A context-aware distance measure.} 
The exponential decay function decides the rate at which the attention weights decay as the distance between the current time index and the previous time indices increase. 
A straightforward way to define the distance between two time indices is their absolute value difference, i.e., $d(t,\tau)= {|t - \tau|}$. However, this distance is not context-aware and ignores the practice history of each learner. For example, consider the two following sequences of concepts a learner practiced on:
\begin{eqnarray*}
{\mbox{Venn Diagram (VD)}}_1,{\mbox{VD}}_2,\cdots,{\mbox{VD}}_8, {\mbox{Prime Numbers (PN)}}_9, {\mbox{PN}}_{10}
\end{eqnarray*}
and
\begin{eqnarray*}
{\mbox{PN}}_1, {\mbox{VD}}_2,{\mbox{VD}}_3, \cdots, {\mbox{VD}}_9,
{\mbox{PN}}_{10},
\end{eqnarray*}
where the notation ``$VD_2$'' means that the learner practiced the concept of Venn Diagram at time step~$2$. In this example, the learner answers a question on Prime Numbers at $t=10$, i.e., the current time index, in both of these sequences, but the most recent past practice on Prime Numbers comes at different time indices. Since the concepts of Venn Diagram and Prime Numbers are not closely related, the learner's previous practices on Prime Numbers is more relevant to us when predicting their answer to the current practice question than recent practices on Venn Diagram. In this case, with the straightforward absolute value difference, an exponential decay curve will significantly reduce the attention weight assigned to the practice on Prime Numbers at $t=1$.

Therefore, we propose the following context-aware distance measure between time steps $d(t,\tau)$ with $\tau \leq t$ for the exponential decay mechanism (in the encoders):
\begin{align*}
d(t, \tau) &= |t-\tau| \cdot \sum_{t'=\tau+1}^{t}\gamma_{t, t'}, \\ \gamma_{t,t'} &= \frac{\exp{ (\frac{\bq_t^{\intercal} \bk_{t'}}{\sqrt{D_k}} )}}{\sum_{1 \leq \tau'\leq t} \exp{ (\frac{\bq_t^{\intercal} \bk_{\tau'}}{\sqrt{D_k}})}}, \, \forall t'\leq t.
\end{align*}
For the knowledge retriever, we replace $\tau'\leq t$ with $\tau<t$ and $t' \leq t$ with $t' <t$ correspondingly. 
In other words, this context-aware distance measure uses another softmax function to adjust the distance between consecutive time indices according to how the concept practiced in the past is related to the current concept.
In practice, in each iteration during model training, we use the current AKT model parameters to compute the modified distance measure and fix it; we do not pass gradients through the distance measure. 

\textbf{Multi-head attention and sub-layers.} 
We also incorporate multi-head attention which is effective in attending to past positions at multiple time scales \cite{transformer}. Therefore, we use $H$ independent attention heads where every head has its own decay rate $\theta$, concatenate the final output into a $(D_v \cdot H) \times 1$ vector and pass it to the next layer. 
This model design enables AKT to summarize past learner performance at multiple time scales, which bears some similarities to the multiple time windows in the multiscale context, DASH, and DAS3H models \cite{mcm,dash,das3h}. 
We also use several sub-layers, including one for layer normalization \cite{layer}, one for dropout \cite{dropout}, a fully-connected feedforward layer, and a residual connection layer \cite{resnet} in each encoder and the knowledge retriever.

\subsection{Response Prediction}
\label{sec:rasch}

The last component of the AKT method predicts the learner's response to the current question. The input to the prediction model is a vector that concatenates both the retrieved knowledge (the knowledge retriever output $\bh_t$) and the current question embedding ${\bx}_t$; this input goes through another fully-connected network 
before finally going through the sigmoid function \cite{dlbook} to generate the predicted probability $\hat{r}_t \in [0,1]$ that the learner answers the current question correctly. All learnable parameters in the entire AKT method are trained in end-to-end fashion by minimizing the binary cross-entropy loss of all learner responses, i.e., \[\ell =\textstyle\sum_{i} \textstyle\sum_t -(r_t^i \log \hat{r}_t^i + (1- r_t^i) \log (1- \hat{r}_t^i)).\]  

\subsection{Rasch Model-Based Embeddings} 
As we discussed above, existing KT methods use concepts to index questions, i.e., setting $q_t = c_t$. 
This setup is necessary due to data sparsity. 
Let $Q$ denote the total number of questions and $L$ denote the number of learners. In most real-world learner response datasets, the number of learner responses is comparable to $CL$ and much less than $QL$ since many questions are assigned to few learners.  
Therefore, using concepts to index questions is effective in avoiding overparameterization and overfitting. 
However, this basic setup ignores the individual differences among questions covering the same concept, thus limiting the flexibility of KT methods and their potential for personalization. 

We use a classic yet powerful model in psychometrics, the Rasch model (which is also known as the 1PL IRT model) \cite{lordirt,rasch}, to construct raw question and knowledge embeddings. The Rasch model characterizes the probability that a learner answers a question correctly using two scalars: the question's difficulty, and the learner's ability. Despite its simplicity, it has shown to achieve comparable performance to more sophisticated models on learner performance prediction in formal assessments when knowledge is static \cite{db,notdeep}. Specifically, 
we construct the embedding of the question $q_t$ from concept $c_t$ at time step $t$ as 
\begin{align*}
\bx_t = \bc_{c_t} + \mu_{q_t} \cdot \bd_{c_t}, 
\end{align*}
where $\bc_{c_t} \in \mathbb{R}^{D}$ is the embedding of the concept this question covers, and $\bd_{c_t} \in \mathbb{R}^{D}$ is a vector that summarizes the variation in questions covering this concept, and $\mu_{q_t} \in \mathbb{R}$ is a scalar \emph{difficulty} parameter that controls how far this question deviates from the concept it covers. The question-response pairs $(q_t, r_t)$ from concept $c_t$ are extended similarly using the scalar difficulty parameter for each pair:
\begin{align*}
\by_t = \be_{(c_t,r_t)} + \mu_{q_t} \cdot \bff_{(c_t, r_t)}, 
\end{align*}
where $\be_{(c_t,r_t)} \in \mathbb{R}^{D}$ and $\bff_{(c_t,r_t)} \in \mathbb{R}^{D}$ are concept-response embedding and variation vectors.
This model choice reflects our third intuition: \emph{questions labeled as covering the same concept are closely related but have important individual differences that should not be ignored.} This model choice is partly inspired by another work in fusing KT and IRT models \cite{mozerfuse}. 

These Rasch model-based embeddings strike the right balance between modeling individual question differences and avoiding overparameterization. For the question embeddings, the total number of embedding parameters in this model is $2CD + Q$, which is slightly more than that in a model that uses concepts to index questions ($CD$), but much less than that in a model where each question is parameterized individually ($QD$), since $C \ll Q$ and $D \gg 1$. We further define the concept-response embeddings as $ \be_{(c_t,r_t)}=\bc_{c_t}+\mathbf{g}_{r_t}$, where $\mathbf{g}_1$ and $\mathbf{g}_0$ denote the embeddings for correct and incorrect responses (regardless of the concept), respectively. Therefore, we only introduce a total of $(C+2)D+Q$ new embedding parameters instead of $2CD+Q$ new parameters for the concept-response embeddings. We note that our question and question-response embeddings share a group of parameters ($\bc_{c_t}$); this setting is different from existing neural network-based KT methods where the two are independent of each other. 
These compact embedding representations significantly reduce the number of parameters in not only AKT but also some other KT methods, leading to improved performance on future learner performance prediction; see Table~\ref{tab:rasch} for details. 

	\section{Experimental Results}
\label{sec:experiments}
In this section, we detail a series of experiments we conducted to test on several real-world datasets. We evaluate AKT both quantitatively through predicting future learner responses and qualitatively through a series of visualizations and case studies.

 \begin{table}[t]\centering
    %\vspace{-1.0cm}
     \scalebox{.9}{
     \begin{tabular}{llllll }\toprule
         Dataset & Learners & Concepts & Questions & Responses \\
         \midrule
         Statics2011 & $333$ & $1,223$ & - & $189,297$ \\
         ASSISTments2009 & $4,151$ & $110$ & $16,891$ & $325,637$ \\
         ASSISTments2015 & $19,840$ & $100$ & - & $683,801$ \\
         ASSISTments2017 & $1,709$ & $102$ & $3,162$ & $942,816$ \\

         \bottomrule
     \end{tabular}
     }
     %\vspace{-0.1cm}
     \caption{Dataset statistics.}
     \label{tab:main}
    % \vspace{-0.8cm}
 \end{table}

\begin{table*}[t]\centering
    %\vspace{-1.0cm}
    \scalebox{0.97}{
        %\resizebox{\linewidth}{!}{
        \begin{tabular}{l lllllll }\toprule
            Dataset & $\mbox{BKT+}$&   DKT & DKT+ & DKVMN & SAKT & AKT-NR & AKT-R  \\ %&Rasch-AKT
            \midrule
            Statics2011 & $\sim 0.75$ & $ 0.8233\pm 0.0039$ & ${\bf 0.8301\pm 0.0039}$ & $ 0.8195\pm 0.0041$ & $ 0.8029\pm 0.0032$ & ${\it 0.8265\pm 0.0049}$ & \\ 
ASSISTments2009 & $\sim 0.69$ & ${\it 0.817\pm 0.0043}$ & $ 0.8024\pm 0.0045$ & $ 0.8093\pm 0.0044$ & $ 0.752\pm 0.004$ & $ 0.8169\pm 0.0045$ & ${\bf 0.8346\pm 0.0036}$ \\
ASSISTments2015 & & $ 0.731\pm 0.0018$ & ${\it 0.7313\pm 0.0018}$ & $ 0.7276\pm 0.0017$ & $ 0.7212\pm 0.002$ & ${\bf 0.7828\pm 0.0019}$ & \\
ASSISTments2017 & & $ 0.7263\pm 0.0054$ & $ 0.7124\pm 0.0041$ & $ 0.7073\pm 0.0044$ & $ 0.6569\pm 0.0027$ & ${\it 0.7282\pm 0.0037}$ & ${\bf 0.7702\pm 0.0026}$ \\
            
            \bottomrule
        \end{tabular}
    }%}

    \caption{Performance of all KT methods on all datasets in predicting future learner responses. AKT (sometimes significantly) outperforms all baseline methods on all datasets. Best models are bold, second best models are italic.}
    \label{tab:bestversion}
    \vspace{-0.3cm}
\end{table*}

\subsection{Experimental Setup}
%\paragraph{Datasets.} 
\sloppy 
\textbf{Datasets.}
We evaluate the performance of AKT and several baselines on predicting future learner responses using four benchmark datasets: ASSISTments2009,  
ASSISTments2015, 
 ASSISTments2017\footnote{The ASSISTments datasets are retrieved from \\ \href{https://sites.google.com/site/assistmentsdata/home}{https://sites.google.com/site/assistmentsdata/home} and \\ \href{https://sites.google.com/view/assistmentsdatamining/}{https://sites.google.com/view/assistmentsdatamining/}.}, 
and Statics2011.\footnote{The Statics2011 dataset is retrieved from \\ \href{https://pslcdatashop.web.cmu.edu/DatasetInfo?datasetId=507}{https://pslcdatashop.web.cmu.edu/DatasetInfo?datasetId=507}.} %\cite{dkvmn,dkt}.
The ASSISTments datasets were collected from an online tutoring platform; in particular, the ASSISTments2009 dataset has been the standard benchmark for KT methods over the last decade. 
The Statics2011 dataset was collected from a college-level engineering course on statics. 
On all these datasets, we follow a series of standard pre-processing steps in the literature. 
For the ASSISTments2009 dataset, we remove all interactions that are not associated to a named concept. 
For the ASSISTments2015 dataset, we remove all interactions where the ``isCorrect'' field is not 0 or 1. 
We list the numbers of learners, concepts, questions, and question-response pairs in Table~\ref{tab:main}. 
Out of these datasets, only the ASSISTments2009 and ASSISTments2017 datasets contain question IDs; therefore, the Rasch model-based embeddings are only applicable to these two datasets.

%\vspace{-0.1in}
\textbf{Baseline methods and evaluation metric.}
We compare AKT against several baseline KT methods, including BKT+ \cite{bkt}, DKT, DKT+ (which is an improved version of DKT with regularization on prediction consistency \cite{dktplus}), DKVMN \cite{dkvmn}, and the recently proposed self-attentive KT (SAKT) method \cite{sakt}, which uses an attention mechanism that can be viewed as a special case of AKT without context-aware representations of questions and responses and the monotonic attention mechanism. 
We use the area under the receiver operating characteristics curve (AUC) as the metric to evaluate the performance of all KT methods on predicting binary-valued future learner responses to questions.

%\vspace{-0.1in}
\textbf{Training and testing.}
For evaluation purposes, we perform standard k-fold cross-validation (with $k=5$) for all models and all datasets. Thus, for each fold, $20\%$ learners are used as the test set, $20\%$ are used as the validation set, and $60\%$ are used as the training set. For each fold, we use the validation set to perform early stopping and tune the parameters for every KT method.

We truncate learner response sequences that are longer than $200$, following \cite{dkt,dkvmn}, for computational efficiency reasons. If a learner has more than $200$ responses, we break up their entire sequence into multiple shorter sequences. 
We use the Adam optimizer to train all models \cite{adam} with a batch size of $24$ learners to ensure that an entire batch can fit into the memory of our machine (equipped with one NVIDIA Titan X GPU).
We implement all versions of AKT in \texttt{PyTorch}; We also re-implement DKT, DKT+, and SAKT, since including question IDs requires new dataset partitions and leads to new experimental results. 
We use the Xavier parameter initialization method \cite{xavier} for AKT, DKT, DKT+, and SAKT; for DKVMN, we follow their work and use samples from normal distributions to initialize the parameters \cite{dkvmn}. 
We do not re-implement BKT+; its performance on various datasets is taken from \cite{dkvmn}.
For most datasets and most algorithms, one training epoch takes less than 10 seconds. We set the maximum number of epochs to $300$.

\begin{table}[t]\centering
    %\vspace{-1.0cm}
    \scalebox{0.9}{
        %\resizebox{\linewidth}{!}{
        \begin{tabular}{l ll|ll}\toprule
            Dataset & $\mbox{AKT}^{\mbox{raw}}$-NR & $\mbox{AKT-NR}$ & $\mbox{AKT}^{\mbox{raw}}$-R  & AKT-R \\
            \midrule
Statics2011  & $ 0.8253$& ${\bf 0.8265}$&  & \\
ASSISTments2009  & $ 0.8082$& ${\bf 0.8169}$& $ 0.8267$& ${\bf 0.8346}$\\
ASSISTments2015  & $ 0.7332$& ${\bf 0.7828}$&  & \\
ASSISTments2017  & $ 0.7066$& ${\bf 0.7282}$& $ 0.7552$& ${\bf 0.7702}$\\
            \bottomrule
        \end{tabular}
    }%}
    %\vspace{-0.1cm}
    \caption{AKT outperforms its variants that do not use contextual-aware question and response representations.}
    \label{tab:raw}
 %   \vspace{-0.3cm}
\end{table}

\begin{table}[t]\centering
    %\vspace{-1.0cm}
    \scalebox{0.9}{
        %\resizebox{\linewidth}{!}{
        \begin{tabular}{l llllllll}\toprule
            Dataset & SAKT & $\mbox{AKT-NR}^{\mbox{pos}}$ &  $\mbox{AKT-NR}^{\mbox{fixed}}$ & AKT-NR \\
            \midrule
Statics2011  & $ 0.8029$& $ 0.8196$& $ 0.8196$& ${\bf 0.8265}$\\
ASSISTments2009  & $ 0.752$& $ 0.7706$& $ 0.7708$& ${\bf 0.8169}$\\
ASSISTments2015  & $ 0.7212$& $ 0.7271$& $ 0.7272$& ${\bf 0.7828}$\\
ASSISTments2017 & $ 0.6569$& $ 0.672$& $ 0.6722$& ${\bf 0.7282}$\\
            \bottomrule
        \end{tabular}
    }%}
    %\vspace{-0.1cm}
    \caption{AKT significantly outperforms its variants that do not use monotonic attention.}
    \label{tab:monotonic}
  %  \vspace{-0.7cm}
\end{table}

\begin{table*}[t]\centering
    %\vspace{-1.0cm}
    \scalebox{1}{
        %\resizebox{\linewidth}{!}{
        \begin{tabular}{l llllllllll}\toprule
            Dataset & {\small DKT}&   {\small DKT-R} & {\small DKT+}&   {\small DKT+-R}& {\small DKVMN} & {\small DKVMN-R} & {\small SAKT}  &{\small SAKT-R} &{\small AKT-NR}  &{\small AKT-R}\\
            \midrule
{\small ASSISTments2009}  & $ 0.817$& $ 0.8179$& $ 0.8024$& $ 0.8033$& $ 0.8093$& $ 0.8235$& $ 0.752$& $ 0.7784$& $ 0.8169$& ${\bf 0.8346}$\\
{\small ASSISTments2017}  & $ 0.7263$& $ 0.7543$& $ 0.7124$& $ 0.7382$& $ 0.7073$& $ 0.7628$& $ 0.6569$& $ 0.7137$& $ 0.7282$& ${\bf 0.7702}$\\
            \bottomrule
        \end{tabular}
    }%}
%    \vspace{-0.1cm}
    \caption{The Rasch model-based embeddings (sometimes significantly) improve the performance of KT methods.}%\gl{the table header seems confusing. Maybe AKT-R as the last column name?}}
    \label{tab:rasch}
  %  \vspace{-0.4cm}
\end{table*}

\subsection{Results and Discussion}
Table~\ref{tab:bestversion} lists the performance of all KT methods across all datasets on predicting future learner responses; we report the averages as well as the standard deviations across five test folds.  AKT-R and AKT-NR represent variants of the AKT model with and without the Rasch model-based embeddings, respectively.
We see that AKT (sometimes significantly) outperforms other KT methods on the ASSISTments datasets while DKT+ marginally outperforms AKT on the smallest Statics2011 dataset.  
In general, AKT performs better on larger datasets; this result suggests that attention mechanisms are more flexible than recurrent neural networks and are thus more capable of capturing the rich information contained in large-scale real-world learner response datasets. 
On the ASSISTments2015 and ASSISTments2017 datasets, AKT-NR improves the AUC by $6\%$ and $1\%$ over the closest baseline. It performs on-par with the best-performing baseline on the Statics2011 and ASSISTments2009 datasets. 
More importantly, on the ASSISTments2009 and 2017 datasets with question IDs, AKT-R significantly outperforms other KT methods, by $2\%$ and $6\%$ over the closest baseline, respectively. 
We note that DKT outperforms the more advanced DKVMN method in our implementation. While we are able to replicate the performance of DKVMN using the same experimental setting \cite{dkvmn}, we found that DKT performs much better than previously reported in that work. DKT+ performs on-par with DKT, with minor improvements on the Statics2011 dataset. 
We also observe that the RNN-based model, DKT, outperforms SAKT on all datasets.

\textbf{Ablation study.}
In order to justify the three key innovations in the AKT method, context-aware representations of questions and responses, the monotonic attention mechanism, and the Rasch model-based embeddings, we perform three additional ablation experiments comparing several variants of the AKT method. 
The first experiment compares AKT-NR and AKT-R using context-aware question and response representations (with the question and knowledge encoders) with two variants $\mbox{AKT}^{\mbox{raw}}$-NR and $\mbox{AKT}^{\mbox{raw}}$-R; In these variants, we use raw question and response embeddings as their representations instead of the context-aware representations (i.e., without passing them through the encoders).
The second experiment compares AKT-NR against several variants without the monotonic attention mechanism. These variants include $\mbox{AKT-NR}^{\mbox{pos}}$, which uses (learnable) positional encoding to capture temporal dependencies in learner response data and $\mbox{AKT-NR}^{\mbox{fixed}}$, which uses (fixed) positional encoding using different frequencies of sine and cosine functions \cite{transformer}. 
The third experiment compares AKT-R with AKT-NR, DKT, DKT-R, DKT+, DKT+-R, DKVMN, DKVMN-R, SAKT, and SAKT-R on the ASSISTments2009 and 2017 datasets where question IDs are available; DKT-R, DKT+-R, DKVMN-R, and SAKT-R refer to the DKT, DKT+, DKVMN, and SAKT methods augmented with the Rasch model-based embeddings as input, respectively.

Table~\ref{tab:raw} shows the results (only averages and not standard deviations across test folds, due to spatial constraints) of the first ablation experiment for the context-aware representations (i.e., the question and knowledge encoders). On all datasets %except the smallest Statics2011 dataset,
AKT-R and AKT-NR outperform their counterparts, $\mbox{AKT}^{\mbox{raw}}$-NR and $\mbox{AKT}^{\mbox{raw}}$-R, which use only a single self-attention mechanism with exponential decay (i.e., the knowledge retriever). 
These results suggest that our context-aware representations of questions and responses are effective at summarizing each learner's  practice history. 

Table~\ref{tab:monotonic} shows the results of the second ablation experiment for the monotonic attention mechanism. We see that AKT-NR significantly outperforms other attention mechanisms using positional embeddings, including SAKT, by about $1\%$ to $6\%$ on all datasets. 
We postulate that the reason for this result is that unlike in language tasks where strong long-distance dependencies between words are more common, the dependence of future learner performance on the past is restricted to a much shorter time window. 
Therefore, using multi-head attention with different exponential decay rates in the attention weights can effectively capture short-term dependencies on the past at different time scales.

Table~\ref{tab:rasch} shows the results of the third ablation experiment for the Rasch model-based embeddings on the two ASSISTments datasets where question IDs are available. All baseline KT methods with the added Rasch model-based embeddings outperform their regular versions, especially on the ASSISTments2017 dataset. 
These results confirm our intuition that treating all questions covering the same concept as a single question is problematic; individual differences among these questions should not be ignored as long as overparameterization can be avoided.

\textbf{Remark.} 
Our standard experimental setting follows that used in \cite{dkvmn,dkt}. In this setting, for questions tagged with multiple concepts (in the ASSISTments2009 dataset), a single learner response is repeated multiple times, one for each concept. Other works used different experimental settings for these questions; In \cite{notdeep}, the authors removed such questions and as a result, DKT's performance dropped to 0.71. In \cite{goingdeeper}, the authors built new concepts for each combination of co-occurring single concepts and as a result, DKT's performance dropped to 0.73. Therefore, we also use an alternative experimental setting on the ASSISTments2009 dataset. 
For a question tagged with multiple concepts, we average the corresponding concept embeddings and use them as both input embeddings and for response prediction. Table~\ref{tab:alternate-skills} lists the performance of all KT methods on the ASSISTments2009 dataset under this setting.
DKT's performance dropped to 0.76 using average embeddings, faring better than settings under \cite{goingdeeper,notdeep}. We observe similar performance drops compared to our standard experimental setting for all KT methods, while AKT-R still comfortably outperforms all baselines. 

\begin{table}[t]\centering
    %\vspace{-1.0cm}
    \scalebox{0.9}{
        %\resizebox{\linewidth}{!}{
        \begin{tabular}{l llllllll}\toprule
DKT & DKT+ & DKVMN & SAKT & AKT-NR & AKT-R\\
\midrule
$ 0.7616$& $ 0.7552$& $ 0.7556$& $ 0.7432$& ${\it 0.7627}$& ${\bf 0.7866}$ \\
\bottomrule
\end{tabular}
    }%}
    %\vspace{-0.1cm}
    \caption{AKT still outperforms other KT methods on the ASSISTments2009 dataset under an alternative experimental setting for questions tagged with multiple concepts.}
    \label{tab:alternate-skills}
   % \vspace{-0.4cm}
\end{table}

\subsection{Visualizing Learned AKT Parameters}

\begin{figure}[t]
\vspace{-0.5cm}
\centering
\subfigure[]{

\includegraphics[width=1\columnwidth]{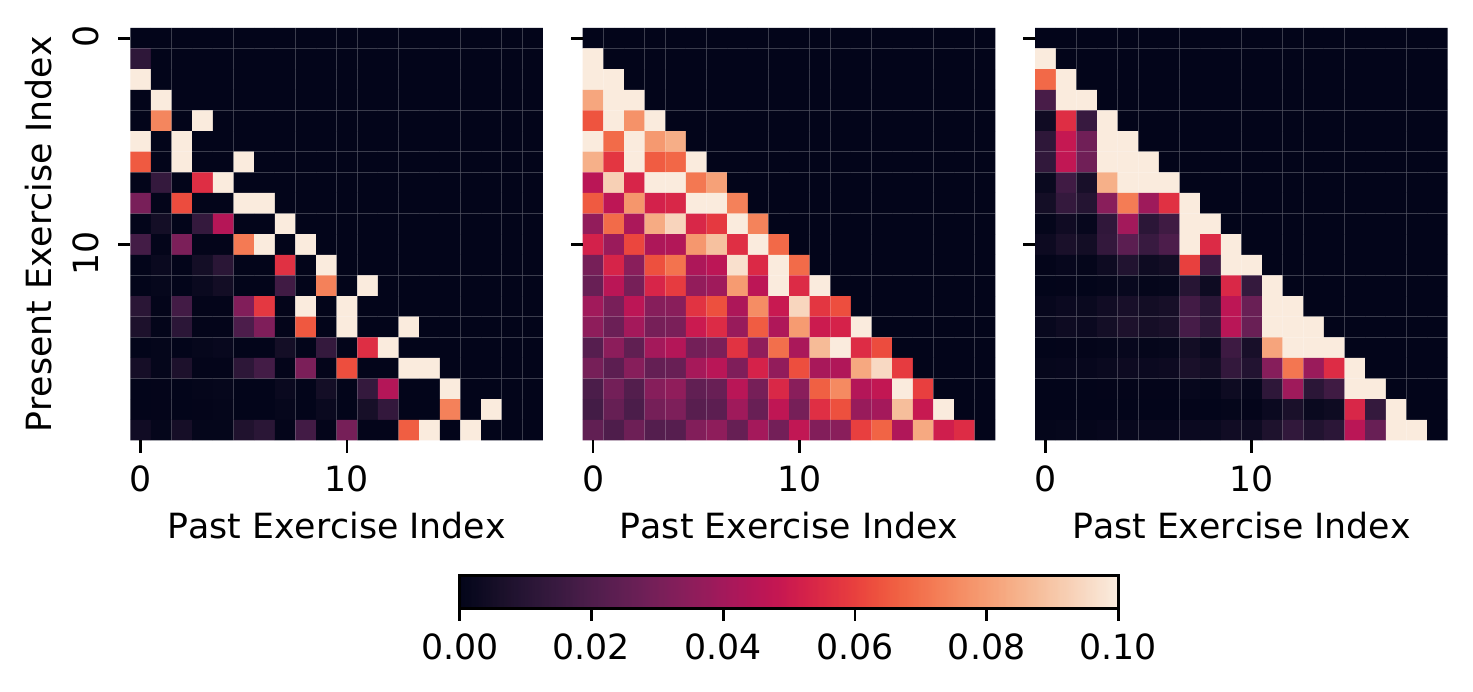}
\label{fig:heatmapviz}
} %\\%[0.1cm]
%
%\vspace{-0.5cm}
\subfigure[]{
\includegraphics[width=1\columnwidth]{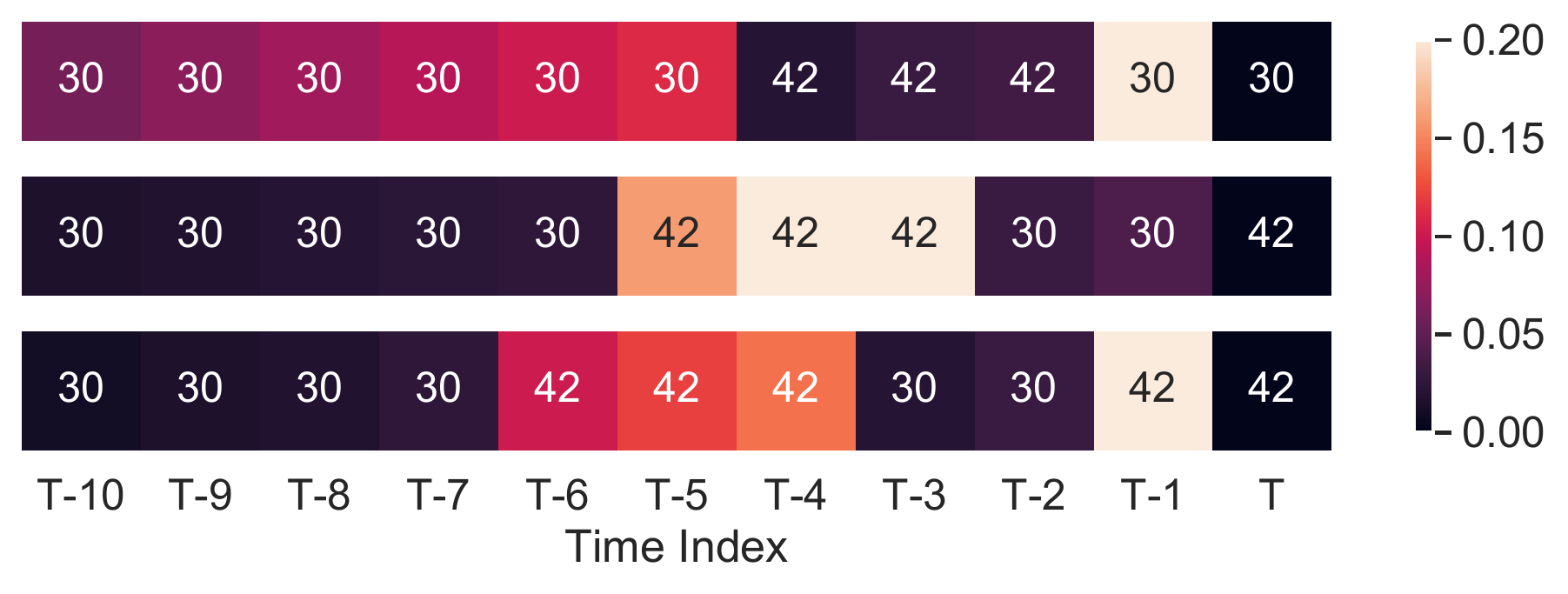}
\label{fig:ktviz}
}
 %
%\vspace{-0.2cm}
\caption{Visualizations of \subref{fig:heatmapviz} attention weights in the decoder of AKT for three attention heads and \subref{fig:ktviz} attention weights for three consecutive practice questions for a learner. Concept similarity and recency are key factor that control the attention weights.}
\label{fig:attn}
\vspace{-0.2cm}
\end{figure}

\textbf{Monotonic attention.} 
Figure~\ref{fig:attn} shows the interpretability offered by AKT's monotonic attention mechanism using the ASSISTments2009 dataset. 
Figure~\ref{fig:heatmapviz} visualizes the attention weights in the knowledge retriever for one learner as an example; we plot the attention weights used to predict their performance on 20 consecutive practice questions across three attention heads. 
We see that each attention head operates on its own time scale: they all have attention windows of different widths. For example, the second head is capable of attending to the entire past, up to 20 time steps (in this example); on the contrary, the third head can only attend to the immediate past and focuses primarily on the last 3-5 time steps. 
This observation suggests that some questions and responses in the past contain information that is highly predictive of the learner's response to the current question; this information can be effectively captured by multiple attention heads with different decay rates. 

Figure~\ref{fig:ktviz} visualizes the normalized attention weights in the knowledge retriever for a single learner for three consecutive time steps. In the top row, the learner is responding to a question on Concept 30 at time $T$ after practicing this concept from $T-10$ to $T-5$, then taking a break to practice on Concept 42, before coming back to Concept 30 at time $T-1$. We see that AKT predicts their response to the current question by focusing more on previous practices on this concept (both in the immediate past and further back) than practices on another concept also in the immediate past. 
In the middle row, the learner switches to practicing on Concept 42 again. Again, AKT learns to focus on past practices on the same concept rather than the immediate past on a different concept at times $T-2$ and $T-1$.  
In the bottom row, the learner practices on Concept 42 for the second consecutive time, and AKT shows a similar focus pattern to that in the top row, with the roles of Concepts 30 and 42 swapped. 
These observations suggest that AKT's monotonic attention mechanism has the potential to provide feedback to teachers by linking a learner's current response to their responses in the past; this information may enable teachers to select certain questions that they have practiced for them to re-practice and clear misconceptions before moving on. We also note that AKT, using a data-driven approach, learns these attention patterns that match hand-crafted features in existing KT methods (e.g., the number of total attempts and correct attempts on this concept) \cite{pfa,dash}.

\begin{figure}[t]
    %\vspace{-0.2cm}
    \centering
    %\subfigure[]{
        
        \includegraphics[width=0.98\columnwidth]{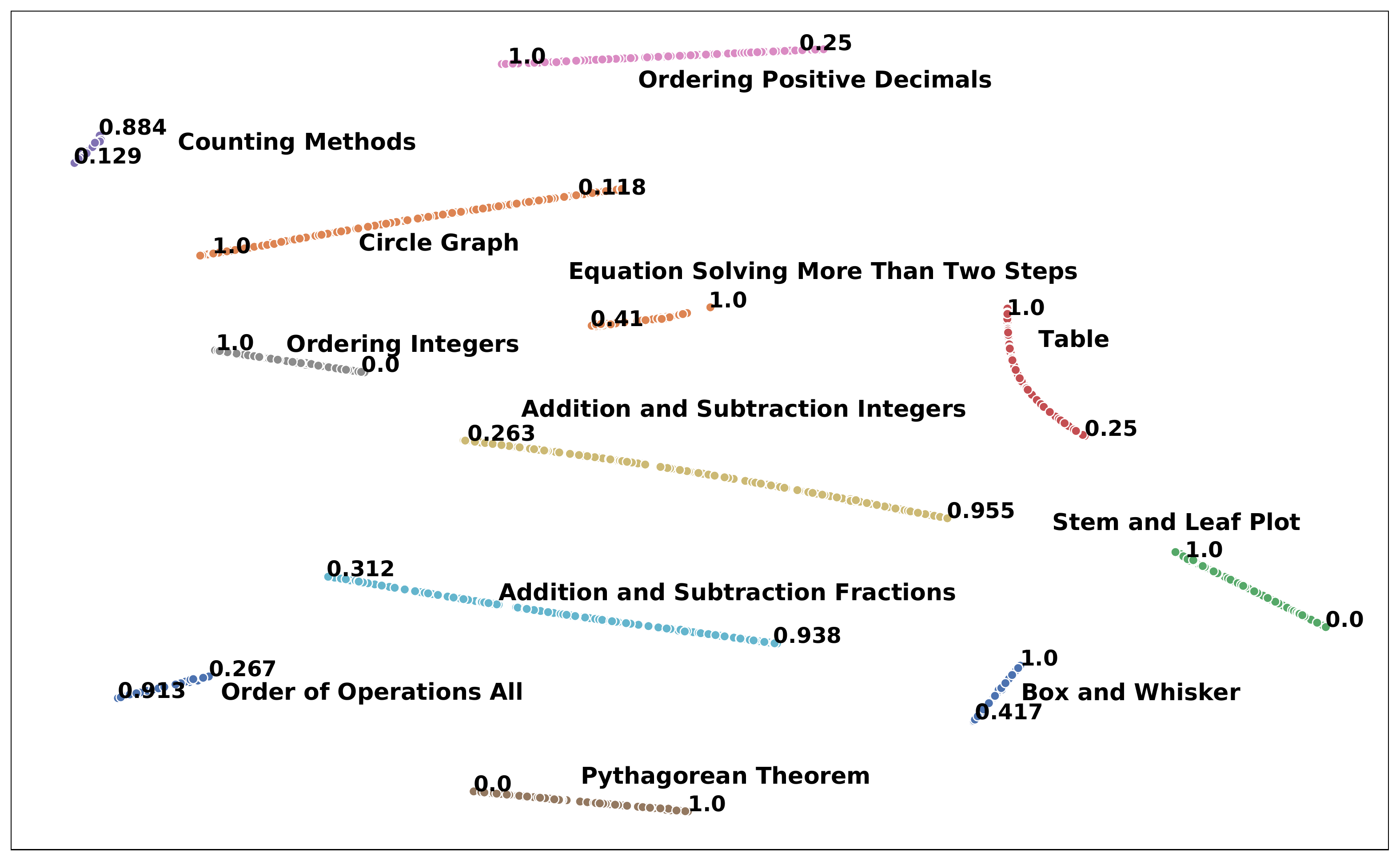}
    \caption{Visualization of learned question embeddings with fraction of correct responses among learners for selected concepts. %\ml{we should probably split this in two - a figure for the embeddings alone, and a table for the difficulties (combine this with table 6, maybe use this to replace the middle concept}
    }
    \label{fig:rasch}
\end{figure}

\textbf{Rasch model-based embeddings.}
Figure~\ref{fig:rasch} visualizes the learned Rasch model-based question embeddings for several concepts using t-SNE \cite{tsne} using the ASSISTments2009 dataset, together with their empirical difficulties for selected questions (portions of correct responses across learners). We also highlight the hardest and easiest question for each concept based on their empirical difficulties. We see that questions on the same concept form a curve and are ordered by their difficulty levels: for the majority of concepts, questions on one end of the line segment are easy while questions on the other end are hard.
This result confirms our intuition that questions from the same concept are not identical but closely related to each other; this relationship can be well-captured by the Rasch model using its difficulty parameter. 

\begin{table}[h]\centering
    %\vspace{-1.0cm}
    \scalebox{0.9}{
        %\resizebox{\linewidth}{!}{
        \begin{tabular}{L{1.75cm}|C{5.25cm}|C{1.1cm}}\toprule
            Concept &Question & $\mu_q$\\ \hline
    & From the list of numbers below, which number is the largest? 
            6.7,        6.4,         3.4,        5.1 & -0.0397\\ \cline{2-3}
\multirow{3}{1.5cm}{Ordering Positive Decimals} &            Which of following decimals is the smallest?
            0.107,           0.1889,               0.12,         0.11582 & 0.0090\\ \cline{2-3}
            
            & Arrange these numbers from least to greatest.
            -1/4, 12.1, -1.4, -4/4 & 0.0279\\ \hline
        &    Steve has a marble jar, that he likes to randomly select marbles from it to play with. The jar has 5 orange marbles and 5 purple marbles. What is the probability that Steve gets an orange marble from the jar?  & $-0.0515$\\ \cline{2-3}
\multirow{3}{1.5cm}{Probability of a Single Event} &    A bag contains 8 red, 5 green, and 7 blue popsicles. John is going to draw out a popsicle without looking in the bag. What is the probability that he will draw either a green or a blue popsicle? &  $0.0088$ \\ \cline{2-3}
&            A card is selected at random from a standard deck  of 52 cards.  Find the probability of choosing a club or an ace card. Enter your answer as a fraction. & $0.0548$ \\\hline 
            
&            Convert     7/7    into a percent.   & -0.0540\\ \cline{2-3}

\multirow{3}{2cm}{Conversion of Fractions} &            Convert     8/4    into a percent. & 0.0038 \\ \cline{2-3}
            
&            Convert     9/8    into a percent. & 0.0529 \\ [.4cm]
            
            %\midrule
            \bottomrule
        \end{tabular}
    }%}
    %\vspace{0.1cm}
    \caption{Question text and learned difficulty parameters ($\mu_q$) for selected questions on three concepts. Learned difficulty levels match our intuition on the difficulty of these questions.}
    \label{tab:raschq}
    \vspace{-0.3cm}
\end{table}
%\gl{slightly modified this part}
Table~\ref{tab:raschq} lists sample questions each for three different concepts, ``Ordering Positive Decimals'', ``Probability of a Single Event'', and ``Conversion of Fractions to Percents'', and their learned difficulty parameters. We show three questions for each concept: an easy one, an average one, and a hard one. Using the ``Probability of a Single Event'' concept as an example, the learned difficulty parameter values ($\mu_q$) are $-0.0515$  for the easy one, $0.0088$ for the average one, and $0.0548$ for the hard one. These learned difficulty levels match our understanding about the difficulty levels of these questions. 

These results suggest that AKT has the potential to be applied in real-world educational settings. Using the estimated difficulty parameters, a computerized learning platform can either i) automatically select questions with appropriate difficulty levels for each learner given their past responses, or ii) support teachers to adjust course plans by providing them feedback on question difficulty levels learned from real data. Therefore, AKT improves over existing KT methods by not only providing state-of-the-art predictive performance but also exhibiting interpretability and potential for personalized learning. 

	\section{Conclusions and Future Work}
In this paper, we have proposed attentive knowledge tracing, a new method for knowledge tracing that relies fully on attention networks. Our method improves upon existing knowledge tracing methods by building context-aware representations of questions and responses, using a monotonic attention mechanism to summarize past learner performance in the right time scale, and by using the Rasch model to capture individual differences among questions covering the same concept. Experimental results on a series of benchmark real-world learner response datasets show that our method outperforms state-of-the-art KT methods and exhibit excellent interpretability. 
Avenues of future work include i) incorporating question text to further enhance the interpretability of question and concept embeddings and ii) testing whether our method can improve prediction performance on language learning datasets where memory decay occurs \cite{reddy}.

	%\section*{References}
	
	%\medskip
	%\balance
	
	%\newpage
	
	\bibliographystyle{ACM-Reference-Format}
	\bibliography{sparfaclustbib,attnbib}


\begin{thebibliography}{36}

%%% the Web address.

\ifx \showCODEN    \undefined \def \showCODEN     #1{\unskip}     \fi
\ifx \showDOI      \undefined \def \showDOI       #1{#1}\fi
\ifx \showISBNx    \undefined \def \showISBNx     #1{\unskip}     \fi
\ifx \showISBNxiii \undefined \def \showISBNxiii  #1{\unskip}     \fi
\ifx \showISSN     \undefined \def \showISSN      #1{\unskip}     \fi
\ifx \showLCCN     \undefined \def \showLCCN      #1{\unskip}     \fi
\ifx \shownote     \undefined \def \shownote      #1{#1}          \fi
\ifx \showarticletitle \undefined \def \showarticletitle #1{#1}   \fi
\ifx \showURL      \undefined \def \showURL       {\relax}        \fi
% invisible to TeX
\providecommand\bibfield[2]{#2}
\providecommand\bibinfo[2]{#2}
\providecommand\natexlab[1]{#1}
\providecommand\showeprint[2][]{arXiv:#2}

\bibitem[\protect\citeauthoryear{Cen, Koedinger, and Junker}{Cen
  et~al\mbox{.}}{2006}]%
        {lfa}
\bibfield{author}{\bibinfo{person}{Hao Cen}, \bibinfo{person}{Kenneth
  Koedinger}, {and} \bibinfo{person}{Brian Junker}.}
  \bibinfo{year}{2006}\natexlab{}.
\newblock \showarticletitle{Learning factors analysis--{A} general method for
  cognitive model evaluation and improvement}. In
  \bibinfo{booktitle}{\emph{Proc. International Conference on Intelligent
  Tutoring Systems}}. \bibinfo{pages}{164--175}.
\newblock


\bibitem[\protect\citeauthoryear{Choffin, Popineau, Bourda, and Vie}{Choffin
  et~al\mbox{.}}{2019}]%
        {das3h}
\bibfield{author}{\bibinfo{person}{Beno{\^\i}t Choffin},
  \bibinfo{person}{Fabrice Popineau}, \bibinfo{person}{Yolaine Bourda}, {and}
  \bibinfo{person}{Jill-J{\^e}nn Vie}.} \bibinfo{year}{2019}\natexlab{}.
\newblock \showarticletitle{DAS3H: {M}odeling student learning and forgetting
  for pptimally scheduling distributed practice of skills}. In
  \bibinfo{booktitle}{\emph{Proc. International Conference on Educational Data
  Mining}}. \bibinfo{pages}{29--38}.
\newblock


\bibitem[\protect\citeauthoryear{Corbett and Anderson}{Corbett and
  Anderson}{1994}]%
        {kt}
\bibfield{author}{\bibinfo{person}{Albert Corbett} {and} \bibinfo{person}{John
  Anderson}.} \bibinfo{year}{1994}\natexlab{}.
\newblock \showarticletitle{Knowledge tracing: {M}odeling the acquisition of
  procedural knowledge}.
\newblock \bibinfo{journal}{\emph{User Modeling and User-adapted Interaction}}
  \bibinfo{volume}{4}, \bibinfo{number}{4} (\bibinfo{date}{Dec.}
  \bibinfo{year}{1994}), \bibinfo{pages}{253--278}.
\newblock


\bibitem[\protect\citeauthoryear{Glorot and Bengio}{Glorot and Bengio}{2010}]%
        {xavier}
\bibfield{author}{\bibinfo{person}{Xavier Glorot} {and} \bibinfo{person}{Yoshua
  Bengio}.} \bibinfo{year}{2010}\natexlab{}.
\newblock \showarticletitle{Understanding the difficulty of training deep
  feedforward neural networks}. In \bibinfo{booktitle}{\emph{Proc.
  International Conference on Artificial Intelligence and Statistics}}.
  \bibinfo{pages}{249--256}.
\newblock


\bibitem[\protect\citeauthoryear{Goodfellow, Bengio, and Courville}{Goodfellow
  et~al\mbox{.}}{2016}]%
        {dlbook}
\bibfield{author}{\bibinfo{person}{Ian Goodfellow}, \bibinfo{person}{Yoshua
  Bengio}, {and} \bibinfo{person}{Aaron Courville}.}
  \bibinfo{year}{2016}\natexlab{}.
\newblock \bibinfo{booktitle}{\emph{Deep Learning}}.
\newblock \bibinfo{publisher}{MIT Press}.
\newblock


\bibitem[\protect\citeauthoryear{He, Zhang, Ren, and Sun}{He
  et~al\mbox{.}}{2016}]%
        {resnet}
\bibfield{author}{\bibinfo{person}{Kaiming He}, \bibinfo{person}{Xiangyu
  Zhang}, \bibinfo{person}{Shaoqing Ren}, {and} \bibinfo{person}{Jian Sun}.}
  \bibinfo{year}{2016}\natexlab{}.
\newblock \showarticletitle{Deep residual learning for image recognition}. In
  \bibinfo{booktitle}{\emph{Proc. IEEE Conference on Computer Vision and
  Pattern Recognition}}. \bibinfo{pages}{770--778}.
\newblock


\bibitem[\protect\citeauthoryear{Hochreiter and Schmidhuber}{Hochreiter and
  Schmidhuber}{1997}]%
        {lstm}
\bibfield{author}{\bibinfo{person}{Sepp Hochreiter} {and}
  \bibinfo{person}{J{\"{u}}rgen Schmidhuber}.} \bibinfo{year}{1997}\natexlab{}.
\newblock \showarticletitle{Long short-term memory}.
\newblock \bibinfo{journal}{\emph{Neural Computation}} \bibinfo{volume}{9},
  \bibinfo{number}{8} (\bibinfo{date}{Nov.} \bibinfo{year}{1997}),
  \bibinfo{pages}{1735--1780}.
\newblock


\bibitem[\protect\citeauthoryear{Khajah, Huang, Gonz{\'a}lez-Brenes, Mozer, and
  Brusilovsky}{Khajah et~al\mbox{.}}{2014}]%
        {mozerfuse}
\bibfield{author}{\bibinfo{person}{MM Khajah}, \bibinfo{person}{Y Huang},
  \bibinfo{person}{JP Gonz{\'a}lez-Brenes}, \bibinfo{person}{MC Mozer}, {and}
  \bibinfo{person}{P Brusilovsky}.} \bibinfo{year}{2014}\natexlab{}.
\newblock \showarticletitle{Integrating knowledge tracing and item response
  theory: {A} tale of two frameworks}. In \bibinfo{booktitle}{\emph{Proc.
  International Workshop on Personalization Approaches in Learning
  Environments}}, Vol.~\bibinfo{volume}{1181}. \bibinfo{pages}{7--15}.
\newblock


\bibitem[\protect\citeauthoryear{Khajah, Lindsey, and Mozer}{Khajah
  et~al\mbox{.}}{2016}]%
        {howdeep}
\bibfield{author}{\bibinfo{person}{Mohammad Khajah}, \bibinfo{person}{Robert
  Lindsey}, {and} \bibinfo{person}{Michael Mozer}.}
  \bibinfo{year}{2016}\natexlab{}.
\newblock \showarticletitle{How deep is knowledge tracing?}. In
  \bibinfo{booktitle}{\emph{Proc. International Conference on Educational Data
  Mining}}. \bibinfo{pages}{94--101}.
\newblock


\bibitem[\protect\citeauthoryear{Kingma and Ba}{Kingma and Ba}{2015}]%
        {adam}
\bibfield{author}{\bibinfo{person}{Diederik~P Kingma} {and}
  \bibinfo{person}{Jimmy Ba}.} \bibinfo{year}{2015}\natexlab{}.
\newblock \showarticletitle{Adam: {A} method for stochastic optimization}. In
  \bibinfo{booktitle}{\emph{Proc. International Conference on Learning
  Representations}}.
\newblock


\bibitem[\protect\citeauthoryear{Lan and Baraniuk}{Lan and Baraniuk}{2016}]%
        {banditslan}
\bibfield{author}{\bibinfo{person}{Andrew Lan} {and} \bibinfo{person}{Richard
  Baraniuk}.} \bibinfo{year}{2016}\natexlab{}.
\newblock \showarticletitle{A Contextual Bandits Framework for Personalized
  Learning Action Selection}. In \bibinfo{booktitle}{\emph{Proc. International
  Conference on Educational Data Mining}}. \bibinfo{pages}{424--429}.
\newblock


\bibitem[\protect\citeauthoryear{Lan, Goldstein, Baraniuk, and Studer}{Lan
  et~al\mbox{.}}{2016}]%
        {db}
\bibfield{author}{\bibinfo{person}{Andrew Lan}, \bibinfo{person}{Tom
  Goldstein}, \bibinfo{person}{Richard Baraniuk}, {and}
  \bibinfo{person}{Christoph Studer}.} \bibinfo{year}{2016}\natexlab{}.
\newblock \showarticletitle{Dealbreaker: {A} nonlinear latent variable model
  for educational data}. In \bibinfo{booktitle}{\emph{Proc. International
  Conference on Machine Learning}}. \bibinfo{pages}{266--275}.
\newblock


\bibitem[\protect\citeauthoryear{Lan, Studer, and Baraniuk}{Lan
  et~al\mbox{.}}{2014}]%
        {tracekdd}
\bibfield{author}{\bibinfo{person}{Andrew Lan}, \bibinfo{person}{Christoph
  Studer}, {and} \bibinfo{person}{Richard Baraniuk}.}
  \bibinfo{year}{2014}\natexlab{}.
\newblock \showarticletitle{Time-varying learning and content analytics via
  sparse factor analysis}. In \bibinfo{booktitle}{\emph{Proc. ACM SIGKDD
  International Conference on Knowledge Discovery and Data Mining}}.
  \bibinfo{pages}{452--461}.
\newblock


\bibitem[\protect\citeauthoryear{Lei~Ba, Kiros, and Hinton}{Lei~Ba
  et~al\mbox{.}}{2016}]%
        {layer}
\bibfield{author}{\bibinfo{person}{Jimmy Lei~Ba}, \bibinfo{person}{Jamie~Ryan
  Kiros}, {and} \bibinfo{person}{Geoffrey~E Hinton}.}
  \bibinfo{year}{2016}\natexlab{}.
\newblock \showarticletitle{Layer normalization}.
\newblock \bibinfo{journal}{\emph{arXiv preprint arXiv:1607.06450}}
  (\bibinfo{date}{July} \bibinfo{year}{2016}).
\newblock


\bibitem[\protect\citeauthoryear{Lindsey, Shroyer, Pashler, and Mozer}{Lindsey
  et~al\mbox{.}}{2014}]%
        {dash}
\bibfield{author}{\bibinfo{person}{Robert Lindsey}, \bibinfo{person}{Jeffery
  Shroyer}, \bibinfo{person}{Harold Pashler}, {and} \bibinfo{person}{Michael
  Mozer}.} \bibinfo{year}{2014}\natexlab{}.
\newblock \showarticletitle{Improving students' long-term knowledge retention
  through personalized review}.
\newblock \bibinfo{journal}{\emph{Psychological Science}} \bibinfo{volume}{25},
  \bibinfo{number}{3} (\bibinfo{date}{Jan.} \bibinfo{year}{2014}),
  \bibinfo{pages}{639--647}.
\newblock


\bibitem[\protect\citeauthoryear{Lord}{Lord}{1980}]%
        {lordirt}
\bibfield{author}{\bibinfo{person}{Frederick Lord}.}
  \bibinfo{year}{1980}\natexlab{}.
\newblock \bibinfo{booktitle}{\emph{Applications of Item Response Theory to
  Practical Testing Problems}}.
\newblock \bibinfo{publisher}{Erlbaum Associates}.
\newblock


\bibitem[\protect\citeauthoryear{Mozer, Kazakov, and Lindsey}{Mozer
  et~al\mbox{.}}{2017}]%
        {moz}
\bibfield{author}{\bibinfo{person}{Michael Mozer}, \bibinfo{person}{Denis
  Kazakov}, {and} \bibinfo{person}{Robert Lindsey}.}
  \bibinfo{year}{2017}\natexlab{}.
\newblock \showarticletitle{Discrete-event continuous-time recurrent nets}.
\newblock \bibinfo{journal}{\emph{arXiv preprint arXiv:1710.04110}}
  (\bibinfo{date}{Oct.} \bibinfo{year}{2017}).
\newblock


\bibitem[\protect\citeauthoryear{Pandey and Karypis}{Pandey and
  Karypis}{2019}]%
        {sakt}
\bibfield{author}{\bibinfo{person}{Shalini Pandey} {and}
  \bibinfo{person}{George Karypis}.} \bibinfo{year}{2019}\natexlab{}.
\newblock \showarticletitle{A self attentive model for knowledge tracing}. In
  \bibinfo{booktitle}{\emph{Proc. International Conference on Educational Data
  Mining}}. \bibinfo{pages}{384--389}.
\newblock


\bibitem[\protect\citeauthoryear{Pardos and Heffernan}{Pardos and
  Heffernan}{2010}]%
        {pardoskt}
\bibfield{author}{\bibinfo{person}{Zachary Pardos} {and} \bibinfo{person}{Neil
  Heffernan}.} \bibinfo{year}{2010}\natexlab{}.
\newblock \showarticletitle{Modeling individualization in a {B}ayesian networks
  implementation of knowledge tracing}. In \bibinfo{booktitle}{\emph{Proc.
  International Conference on User Modeling, Adaptation, and Personalization}}.
  \bibinfo{pages}{255--266}.
\newblock


\bibitem[\protect\citeauthoryear{Pascanu, Mikolov, and Bengio}{Pascanu
  et~al\mbox{.}}{2013}]%
        {maxgradnorm}
\bibfield{author}{\bibinfo{person}{Razvan Pascanu}, \bibinfo{person}{Tomas
  Mikolov}, {and} \bibinfo{person}{Yoshua Bengio}.}
  \bibinfo{year}{2013}\natexlab{}.
\newblock \showarticletitle{On the difficulty of training recurrent neural
  networks}. In \bibinfo{booktitle}{\emph{Proc. International Conference on
  Machine Learning}}. \bibinfo{pages}{1310--1318}.
\newblock


\bibitem[\protect\citeauthoryear{Pashler, Cepeda, Lindsey, Vul, and
  Mozer}{Pashler et~al\mbox{.}}{2009}]%
        {mcm}
\bibfield{author}{\bibinfo{person}{Harold Pashler}, \bibinfo{person}{Nicholas
  Cepeda}, \bibinfo{person}{Robert Lindsey}, \bibinfo{person}{Ed Vul}, {and}
  \bibinfo{person}{Michael Mozer}.} \bibinfo{year}{2009}\natexlab{}.
\newblock \showarticletitle{Predicting the optimal spacing of study: {A}
  multiscale context model of memory}. In \bibinfo{booktitle}{\emph{Proc.
  Conference on Advances in Neural Information Processing Systems}}.
  \bibinfo{pages}{1321--1329}.
\newblock


\bibitem[\protect\citeauthoryear{Pavlik~Jr, Cen, and Koedinger}{Pavlik~Jr
  et~al\mbox{.}}{2009}]%
        {pfa}
\bibfield{author}{\bibinfo{person}{Philip Pavlik~Jr}, \bibinfo{person}{Hao
  Cen}, {and} \bibinfo{person}{Kenneth Koedinger}.}
  \bibinfo{year}{2009}\natexlab{}.
\newblock \showarticletitle{Performance factors analysis--{A} new alternative
  to knowledge tracing}. In \bibinfo{booktitle}{\emph{Proc. International
  Conference on Artificial Intelligence in Education}}.
  \bibinfo{pages}{531--538}.
\newblock


\bibitem[\protect\citeauthoryear{Piech, Bassen, Huang, Ganguli, Sahami, Guibas,
  and Sohl-Dickstein}{Piech et~al\mbox{.}}{2015a}]%
        {dkt}
\bibfield{author}{\bibinfo{person}{Chris Piech}, \bibinfo{person}{Jonathan
  Bassen}, \bibinfo{person}{Jonathan Huang}, \bibinfo{person}{Surya Ganguli},
  \bibinfo{person}{Mehran Sahami}, \bibinfo{person}{Leonidas~J Guibas}, {and}
  \bibinfo{person}{Jascha Sohl-Dickstein}.} \bibinfo{year}{2015}\natexlab{a}.
\newblock \showarticletitle{Deep knowledge tracing}. In
  \bibinfo{booktitle}{\emph{Proc. Conference on Advances in Neural Information
  Processing Systems}}. \bibinfo{pages}{505--513}.
\newblock


\bibitem[\protect\citeauthoryear{Piech, Huang, Nguyen, Phulsuksombati, Sahami,
  and Guibas}{Piech et~al\mbox{.}}{2015b}]%
        {piechfeedback}
\bibfield{author}{\bibinfo{person}{Chris Piech}, \bibinfo{person}{Jonathan
  Huang}, \bibinfo{person}{Andy Nguyen}, \bibinfo{person}{Mike Phulsuksombati},
  \bibinfo{person}{Mehran Sahami}, {and} \bibinfo{person}{Leonidas Guibas}.}
  \bibinfo{year}{2015}\natexlab{b}.
\newblock \showarticletitle{Learning Program Embeddings to Propagate Feedback
  on Student Code}. In \bibinfo{booktitle}{\emph{Proc. International Conference
  on Machine Learning}}. \bibinfo{pages}{1093--1102}.
\newblock


\bibitem[\protect\citeauthoryear{Rasch}{Rasch}{1993}]%
        {rasch}
\bibfield{author}{\bibinfo{person}{Georg Rasch}.}
  \bibinfo{year}{1993}\natexlab{}.
\newblock \bibinfo{booktitle}{\emph{Probabilistic Models for Some Intelligence
  and Attainment Tests}}.
\newblock \bibinfo{publisher}{MESA Press}.
\newblock


\bibitem[\protect\citeauthoryear{Reddy, Labutov, Banerjee, and Joachims}{Reddy
  et~al\mbox{.}}{2016}]%
        {reddy}
\bibfield{author}{\bibinfo{person}{Siddharth Reddy}, \bibinfo{person}{Igor
  Labutov}, \bibinfo{person}{Siddhartha Banerjee}, {and}
  \bibinfo{person}{Thorsten Joachims}.} \bibinfo{year}{2016}\natexlab{}.
\newblock \showarticletitle{Unbounded human learning: {O}ptimal scheduling for
  spaced repetition}. In \bibinfo{booktitle}{\emph{Proc. ACM SIGKDD
  International Conference on Knowledge Discovery and Data Mining}}.
  \bibinfo{pages}{1815--1824}.
\newblock


\bibitem[\protect\citeauthoryear{Srivastava, Hinton, Krizhevsky, Sutskever, and
  Salakhutdinov}{Srivastava et~al\mbox{.}}{2014}]%
        {dropout}
\bibfield{author}{\bibinfo{person}{Nitish Srivastava},
  \bibinfo{person}{Geoffrey Hinton}, \bibinfo{person}{Alex Krizhevsky},
  \bibinfo{person}{Ilya Sutskever}, {and} \bibinfo{person}{Ruslan
  Salakhutdinov}.} \bibinfo{year}{2014}\natexlab{}.
\newblock \showarticletitle{Dropout: {A} simple way to prevent neural networks
  from overfitting}.
\newblock \bibinfo{journal}{\emph{The Journal of Machine Learning Research}}
  \bibinfo{volume}{15}, \bibinfo{number}{1} (\bibinfo{date}{June}
  \bibinfo{year}{2014}), \bibinfo{pages}{1929--1958}.
\newblock


\bibitem[\protect\citeauthoryear{Van Der~Maaten}{Van Der~Maaten}{2014}]%
        {tsne}
\bibfield{author}{\bibinfo{person}{Laurens Van Der~Maaten}.}
  \bibinfo{year}{2014}\natexlab{}.
\newblock \showarticletitle{Accelerating t-SNE using tree-based algorithms}.
\newblock \bibinfo{journal}{\emph{The Journal of Machine Learning Research}}
  \bibinfo{volume}{15}, \bibinfo{number}{1} (\bibinfo{year}{2014}),
  \bibinfo{pages}{3221--3245}.
\newblock


\bibitem[\protect\citeauthoryear{Vaswani, Shazeer, Parmar, Uszkoreit, Jones,
  Gomez, Kaiser, and Polosukhin}{Vaswani et~al\mbox{.}}{2017}]%
        {transformer}
\bibfield{author}{\bibinfo{person}{Ashish Vaswani}, \bibinfo{person}{Noam
  Shazeer}, \bibinfo{person}{Niki Parmar}, \bibinfo{person}{Jakob Uszkoreit},
  \bibinfo{person}{Llion Jones}, \bibinfo{person}{Aidan~N Gomez},
  \bibinfo{person}{{\L}ukasz Kaiser}, {and} \bibinfo{person}{Illia
  Polosukhin}.} \bibinfo{year}{2017}\natexlab{}.
\newblock \showarticletitle{Attention is all you need}. In
  \bibinfo{booktitle}{\emph{Proc. Conference on Advances in Neural Information
  Processing Systems}}. \bibinfo{pages}{5998--6008}.
\newblock


\bibitem[\protect\citeauthoryear{Vie and Kashima}{Vie and Kashima}{2019}]%
        {kfm}
\bibfield{author}{\bibinfo{person}{Jill-J{\^e}nn Vie} {and}
  \bibinfo{person}{Hisashi Kashima}.} \bibinfo{year}{2019}\natexlab{}.
\newblock \showarticletitle{Knowledge tracing machines: {F}actorization
  machines for knowledge tracing}. In \bibinfo{booktitle}{\emph{Proc. AAAI
  Conference on Artificial Intelligence}}, Vol.~\bibinfo{volume}{33}.
  \bibinfo{pages}{750--757}.
\newblock


\bibitem[\protect\citeauthoryear{Wilson, Karklin, Han, and Ekanadham}{Wilson
  et~al\mbox{.}}{2016}]%
        {notdeep}
\bibfield{author}{\bibinfo{person}{Kevin Wilson}, \bibinfo{person}{Yan
  Karklin}, \bibinfo{person}{Bojian Han}, {and} \bibinfo{person}{Chaitanya
  Ekanadham}.} \bibinfo{year}{2016}\natexlab{}.
\newblock \showarticletitle{Back to the basics: {B}ayesian extensions of {IRT}
  outperform neural networks for proficiency estimation}. In
  \bibinfo{booktitle}{\emph{Proc. International Conference on Educational Data
  Mining}}. \bibinfo{pages}{539--544}.
\newblock


\bibitem[\protect\citeauthoryear{Woolf}{Woolf}{2010}]%
        {bevbook}
\bibfield{author}{\bibinfo{person}{Beverly~Park Woolf}.}
  \bibinfo{year}{2010}\natexlab{}.
\newblock \bibinfo{booktitle}{\emph{Building Intelligent Interactive Tutors:
  {S}tudent-centered Strategies for Revolutionizing E-learning}}.
\newblock \bibinfo{publisher}{Morgan Kaufmann}.
\newblock


\bibitem[\protect\citeauthoryear{Xiong, Zhao, Van~Inwegen, and Beck}{Xiong
  et~al\mbox{.}}{2016}]%
        {goingdeeper}
\bibfield{author}{\bibinfo{person}{Xiaolu Xiong}, \bibinfo{person}{Siyuan
  Zhao}, \bibinfo{person}{Eric~G Van~Inwegen}, {and} \bibinfo{person}{Joseph~E
  Beck}.} \bibinfo{year}{2016}\natexlab{}.
\newblock \showarticletitle{Going deeper with deep knowledge tracing}.
\newblock \bibinfo{journal}{\emph{International Educational Data Mining
  Society}} (\bibinfo{year}{2016}).
\newblock


\bibitem[\protect\citeauthoryear{Yeung and Yeung}{Yeung and Yeung}{2018}]%
        {dktplus}
\bibfield{author}{\bibinfo{person}{Chun-Kit Yeung} {and}
  \bibinfo{person}{Dit-Yan Yeung}.} \bibinfo{year}{2018}\natexlab{}.
\newblock \showarticletitle{Addressing two problems in deep knowledge tracing
  via prediction-consistent regularization}. In \bibinfo{booktitle}{\emph{Proc.
  ACM Conference on Learning at Scale}}. ACM, \bibinfo{pages}{5}.
\newblock


\bibitem[\protect\citeauthoryear{Yudelson, Koedinger, and Gordon}{Yudelson
  et~al\mbox{.}}{2013}]%
        {bkt}
\bibfield{author}{\bibinfo{person}{Michael Yudelson}, \bibinfo{person}{Kenneth
  Koedinger}, {and} \bibinfo{person}{Geoffrey Gordon}.}
  \bibinfo{year}{2013}\natexlab{}.
\newblock \showarticletitle{Individualized {B}ayesian knowledge tracing
  models}. In \bibinfo{booktitle}{\emph{Proc. International Conference on
  Artificial Intelligence in Education}}. \bibinfo{pages}{171--180}.
\newblock


\bibitem[\protect\citeauthoryear{Zhang, Shi, King, and Yeung}{Zhang
  et~al\mbox{.}}{2017}]%
        {dkvmn}
\bibfield{author}{\bibinfo{person}{Jiani Zhang}, \bibinfo{person}{Xingjian
  Shi}, \bibinfo{person}{Irwin King}, {and} \bibinfo{person}{Dit-Yan Yeung}.}
  \bibinfo{year}{2017}\natexlab{}.
\newblock \showarticletitle{Dynamic key-value memory networks for knowledge
  tracing}. In \bibinfo{booktitle}{\emph{Proc. International Conference on
  World Wide Web}}. \bibinfo{pages}{765--774}.
\newblock


\end{thebibliography}
	
	%\newpage

%\rule{\textwidth}{0.13cm}
%
\newpage
\begin{center}
\LARGE \bf 
Supplementary Material% to \\ Attentive Knowledge Tracing
\end{center}
%\rule{\textwidth}{0.04cm}

\vspace{1cm}

\textbf{Parameter tuning.} 
For all methods, we use a two-layer, fully-connected network for the response prediction model, with $512$ and $256$ hidden units in the first and second layers, respectively.
For AKT, we fix the number of attention heads to $H=8$ and use the same dimension for the queries, keys, and values for each head in the encoders and the knowledge retriever, i.e., $D_k = D_v  = D/8$ where $D$ is the input embedding dimension. For AKT, we share the query and key embedding layer in the attention mechanism. 
We do not perform grid search over the values of these parameters since the performance of KT methods is insensitive to these parameters. 
For all methods, we use $\{256, 512\}$, $\{256, 512\}$, and $\{0,0.05,0.1,0.15,0.2, 0.25\}$ as values of the input embedding dimension, the hidden state dimension, and the dropout rate for the feedforward network, respectively. We also use $\{1, 10, \infty\}$ as values of the maximum gradient norm for clipping \cite{maxgradnorm} where $\infty$ means no gradient norm clipping. 
For AKT, SAKT, and their variants, we use $\{5\times 10^{-6},10^{-5}, 10^{-4}\}$ as values of the learning rate in the Adam optimizer. 
For DKT, DKT+, and DKVMN, we use $\{10^{-4}, 10^{-3} \}$ as values of the learning rate. 
Additionally, for DKVMN, we use $\{20,50\}$ as values of the memory size parameter, following \cite{dkvmn}.
For DKT+, we use $\{0.1, 0.2\}$, $\{0.3, 1\}$, and $\{3,30\}$ as values of the reconstruction regularization, the $\ell_1$-norm penalty, and the $\ell_2$-norm penalty parameters, respectively, following \cite{dktplus}.
\end{document}